\documentclass[12pt]{article}
\usepackage{arxiv}
\usepackage{authblk}
\usepackage{microtype}
\usepackage{graphicx}
\usepackage{booktabs}
\usepackage{subcaption}
\usepackage{hyperref}

\usepackage[numbers]{natbib}
\usepackage{amsmath}
\usepackage{amssymb}
\usepackage{mathtools}
\usepackage{amsthm}
\usepackage[capitalize,noabbrev]{cleveref}

\theoremstyle{plain}

\theoremstyle{definition}

\theoremstyle{remark}

\newcommand{\be}{\begin{equation}}
\newcommand{\ee}{\end{equation}}
\newcommand{\bea}{\begin{eqnarray}}
\newcommand{\eea}{\end{eqnarray}}

\newcommand{\ccup}[1]{\left\{#1\right\}}

\newcommand{\bup}[1]{\left(#1\right)}

\usepackage[textsize=tiny]{todonotes}

\usepackage[normalem]{ulem} 
\usepackage{soul}

 \newcommand{\mt}{\mbox{{\small MULTITENSOR}}}
  \newcommand{\mtcov}{\mbox{{\small MTCOV}}}
    \newcommand{\bnp}{\mbox{{\small BNP}}}

\begin{document}

\title{How do Probabilistic Graphical Models and Graph Neural Networks Look at Network Data?}

\author[1]{Michela Lapenna}
\author[2,3]{Caterina De Bacco}
\affil[1]{Department of Physics and Astronomy, University of Bologna, Via Irnerio 46, Bologna, 40126, Italy}
\affil[2]{Max Planck Institute for Intelligent Systems, Cyber Valley, Max-Planck-Ring 4, Tübingen, 72076, Germany}
\affil[3]{Faculty of Electrical Engineering, Mathematics and Computer Science, Delft University of Technology, Mekelweg 4, Delft, 2628, The Netherlands}

\date{}
\maketitle

\begin{abstract}
Graphs are a powerful data structure for representing relational data and are widely used to describe complex real-world systems. Probabilistic Graphical Models (PGMs) and Graph Neural Networks (GNNs) can both leverage graph-structured data, but their inherent functioning is different.
The question is how do they compare in capturing the information contained in networked datasets?
We address this objective by solving a link prediction task and we conduct three main experiments, on both synthetic and real networks: one focuses on how PGMs and GNNs handle input features, while the other two investigate their robustness to noisy features and increasing heterophily of the graph. PGMs do not necessarily require features on nodes, while GNNs cannot exploit the network edges alone, and the choice of input features matters. We find that GNNs are outperformed by PGMs when input features are low-dimensional or noisy, mimicking many real scenarios where node attributes might be scalar or noisy.
Then, we find that PGMs are more robust than GNNs when the heterophily of the graph is increased. Finally, to assess performance beyond prediction tasks, we also compare the two frameworks in terms of their computational complexity and interpretability.
\end{abstract}

\noindent\textbf{Keywords:} Probabilistic Graphical Models, Graph Neural Networks, Link Prediction, Interpretability

\medskip

\noindent\textbf{Corresponding authors:} \texttt{michela.lapenna4@unibo.it}, \texttt{c.debacco@tudelft.nl}

\section{Introduction}
\label{intro}

Probabilistic Graphical Models (PGMs) combine principles from graph theory and probability theory to model complex dependencies among random variables \cite{jordan2003introduction}. They offer a structured way to encode prior knowledge and incorporate uncertainty, enabling efficient learning and inference in probabilistic terms. {In their standard framework, they are implemented with linear models and take a limited amount of parameters in input. These models are usually designed from interpretable mechanistic principles guided by domain-knowledge, which makes them inherently interpretable. For instance, the Stochastic Block Model (SMB) \cite{holland, wang_wong, snijders, abbe} is a popular example designed to represent networks where the mechanism guiding edge formation is assumed to be driven by community structure. The learned latent community memberships are parameters that can be directly interpreted to cluster similar nodes, and can be used to solve link prediction tasks.
}

On the other side, Graph Neural Networks (GNNs) are designed to extend traditional neural networks to operate on graph-structured data \cite{scarselli, defferrard, monti}. Their main focus is on solving prediction tasks, taking advantage of the input relational data. They do so by learning representation on nodes by aggregating information among neighborhoods, using message-passing algorithmic updates \cite{veličković2022message}. {They make use of non-linearities and a large number of parameters to learn complex and expressive node representations. These can then be used for tasks such as node classification, graph classification, and link prediction.}

{While their main focus is fundamentally different, they are often used to perform similar tasks. For instance, link prediction is a predictive task that is often adopted in PGMs as a model validation routine. Similarly, the node representations learned by GNNs are often used to cluster nodes and efforts are made to interpret them a posteriori. Hence, in this work we compare them on common grounds aiming at exploring their strengths and weaknesses, by providing insights into the role of the input features and graph structure.}

A fundamental difference between PGMs and GNNs lies in their handling of input features. PGMs do not necessarily require node attributes, as they can rely solely on the graph's edge list. This allows PGMs to function effectively even when node attributes are sparse or unavailable. {However, if node features are available and informative, there is no unique recipe for PGMs to incorporate them and exploit their information. Instead, one has to make ad-hoc modeling extensions to be able to accomodate them in input.   In contrast, GNNs' message passing algorithm explicitly requires in input node features, also when they may not be available or informative}. We thus evaluate how the performance of these approaches is affected by performing experiments where we tune the informativeness, the dimensionality and the availability of input features.

Then, we investigate the robustness of the models when increasing the heterophily of the graph, a structural measure that has been investigated in applications of both modeling approaches and has been shown to increase the difficulty of inference tasks. For this, we consider networks with community structures of different types, from highly cohesive communities to more dispersed and interconnected ones. Since GNNs leverage the structural locality in graphs, we expect that high mixing between communities makes it harder for the model to generalize, as it has been already discussed in \cite{luan,digiovanni,wang,edge_homo,li2022finding,luan2024heterophilic,zhu2024impact}.

To provide a comprehensive evaluation, we conduct experiments on both synthetic and real-world datasets. Synthetic datasets offer controlled environments where we can manipulate network properties and systematically study the models' behaviors under various conditions. Real datasets, on the other hand, typically come equipped with informative feature-engineered vectors in input to the nodes. By testing on both types of datasets, we aim to gain a holistic understanding of each model's strengths and limitations across different contexts.

\section{Related Work}
\label{related}

Our aim is to directly compare Probabilistic Graphical Models (PGMs) and Graph Neural Networks (GNNs) as distinct frameworks for handling network data. This comparison is lacking in the literature, as these two types of approaches usually concern different scientific communities, who adopt distinct languages and distinct model validation criteria, making comparisons hard \cite{blocker2025insights}.

While several studies have explored hybrid approaches combining the strengths of PGMs and GNNs \cite{Mehta2019StochasticBM,boutin2024,wang2023gnninterpreter,yoon,rubio-madrigal2025gnns}, such as leveraging PGMs to enhance the explainability of GNNs \cite{wang2023gnninterpreter} or employing GNNs to improve the scalability and efficiency of PGMs \cite{yoon}, we think that a direct comparison between these two frameworks would provide deeper insights into their respective advantages and limitations. 

In particular, we address the growing recognition in the literature that GNNs often struggle when provided with insufficiently informative input features and when the graph structure is heterophilic, as they cannot fully leverage graph topology alone to achieve robust performance and strongly rely on input features \cite{revisiting, bottleneck}. Moreover, node clustering is often tackled in an unsupervised regime, where node labels are not available as input. In this setting, GNNs have exhibited weaker performance compared to supervised or semi-supervised approaches \cite{tsitsulin2023graph}. Notably, even in a semi-supervised setting, GNNs have been outperformed by the Contextual Stochastic Block Model (CSBM) \cite{csbm} on a node classification task.

\section{Motivation for Link Prediction}
\label{link}

We choose to compare the two frameworks on a link prediction task. Link prediction is a fundamental task in network analysis, where the goal is to infer missing links or predict future connections between nodes in a graph. Link prediction serves our objective of comparing PGMs and GNNs better than node and graph classification because it directly aligns with both models' core capabilities, thus allowing for a fairer comparison. On one hand, differently from a node classification task, link prediction does not require explicit labels on the nodes, making it less reliant on external annotations. {PGMs are in fact often deployed on input data that only contain network edges, with no extra information on nodes, e.g. attributes.} Also, node classification often assumes that network structure is correlated with node labels, to be able to learn from both types of input data. This makes it less general than link prediction, as it relies on observing meaningful node labels, which should not be taken for granted in real datasets, as proved in several works where extra information on nodes was shown not informative in learning from graphs \cite{peel2017ground,newman2016structure,mtcovariate,badalyan2024structure}. On the other hand, graph classification involves predicting a single label for an entire graph. While this is useful for understanding a graph as a whole, it aggregates information across nodes and edges, which can obscure model-specific performance nuances. Also, graph classification is often more domain-specific, focusing on tasks like molecule classification in chemistry or document categorization, while link prediction is a more universal graph problem valid across domains.

In addition, link prediction is often deployed as a model validation task in PGMs, in particular in clustering nodes (e.g. community detection problem) in the absence of ground truth information.
Furthermore, as link prediction often involves a large number of node pairs, it is a valuable benchmark for assessing scalability, efficiency, and the ability to handle sparsity or large graphs. 

\section{Background}
\label{models}

PGMs and GNNs are different approaches to model network structured datasets. Both take as input data $A$, the network adjacency matrix of dimension $N \times N$, where $N$ is the number of nodes. Its entries $A_{ij}$ are the weight of an edge between nodes $i$ and $j$. 
Sometimes, network datasets contain node features $X_i$, $F$-dimensional vectors, where $F$ is the number of features. PGMs may take them in input, but they are not equipped by default to use them. This is in contrast with GNNs, that instead require an input matrix $X$ to train the model.\\
{We imagine a scenario when a practitioner is willing to analyze a network problem and its corresponding dataset, and is agnostic on the model family GNN vs. PGM. Hence, we consider here off-the-shelf implementations of various main models in each of these families, and refrain from using models specifically designed for one main task, e.g. specifically designed for handling heterophilic or disassortative graphs. This also mimics the realistic scenario where the practitioner does not know in advance if their dataset exhibits a specific structure.}

\subsection{PGM models}
\label{pgm}

PGMs describe the dependencies among nodes via the edges of the networks in a probabilistic way. They do so by learning latent variables $\theta$ that control explicitly the probability of observing a given edge weight $A_{ij}$. In this paper we consider three variants of SBM, a model that implements this as:
\begin{equation}
P(A_{ij} | \theta) = P(A_{ij} |f(u_i, u_j, w))\quad,
\end{equation}
where $u_i$ is a $K$-dimensional membership vector (often with positive entries, for interpretability, e.g. for node clustering); $w$ is a $K\times K$ affinity matrix, controlling the density of edges between nodes with different memberships; $f$ is a function combining these parameters into a scalar (e.g. to model the expected value $\mathbb{E}[A_{ij}]$). This is usually a linear function.\\
In this work, we consider three different variants: \mt\ \cite{multitensor} and \mtcov\ \cite{mtcovariate}, two models based on a tensor factorization to parameterize $f$ and that use maximum likelihood estimation for inference; \bnp\ \cite{peixoto2014efficient}, a Bayesian model that parametrizes the $f$ in a non-parametric way and uses Markov chain Monte Carlo sampling for inference. Among these, \mtcov\ is the only model that can also take node attributes in input, thus allowing to assess the impact of using this extra information in addition to the network adjacency matrix. 
{\mt\ and \mtcov\ are mixed-membership, i.e. $u_i$ can contain more than one non-zero entry, while \bnp\ is hard-membership, i.e. $u_i$ is one-hot encoded.}
\subsection{GNN architectures}
\label{gnn}

{Graph Neural Networks (GNNs) \cite{scarselli, defferrard, monti} leverage neural network architectures to learn node embeddings from network structured data. They do so by aggregating the embeddings $h_i^{(\ell)}$ at each layer $\ell$ between node neighbors in a message-passing fashion \cite{veličković2022message}:
\begin{equation}
h_i^{(\ell)} = \phi^{(\ell)}\bup{h_i^{(\ell-1)}, g^{(\ell)}(h^{(\ell-1)}_{\partial i})}\quad,
\end{equation}
where $g^{(\ell)}$ is a permutation-invariant operation (e.g. mean or max) implementing the message-passing routine and $h^{(\ell-1)}_{\partial i} = \ccup{h_j^{(\ell-1)} \,|\, j \in \mathcal{N}_i}$ contains the embeddings of the neighbors $\mathcal{N}_i=\ccup{j \, |  \, A_{ij} >0}$. The embeddings in the first layer are initialized using the input node features $h_i^{(0)}=X_i$. \\
The cost function optimized in training depends on the task. As we focus on link prediction and node clustering, we use binary cross-entropy on the adjacency matrix {(in our experiments we consider binary $A$)}.
This is calculated using the embeddings inferred in the final layer $L$ and applying a scoring function to pairs $(h_i^{(L)}, h_j^{(L)})$, to estimate the probability of existence of an edge $(i,j)$. In this work we combine the embeddings via the dot product, as it is a simple and efficient method, but other choices have been explored in the literature, from concatenation to more complex non-linear methods \cite{paire, automated}. The final embeddings are also the quantities that are often utilized for node clustering, typically by applying K-means \cite{MacQueen, lloyd} on them to assign cluster labels to nodes.\\
The choices of the functions $\phi^{(\ell)}$, $g^{(\ell)}$ differ based on the model. Contrarily to PGMs, they usually entail non-linearities. In our experiments, we employ four distinct architectures: Graph Attention Networks (GAT) \cite{gat}, which use an attention mechanism inside the function $g$; Graph Autoencoders (GAE) \cite{variational} and Graph Variational Autoencoders (VGAE) \cite{variational}, which use an {autoencoder framework} to learn embeddings, {with GAE being a non-probabilistic variant of VGAE}; H\textsubscript{2}GCN~\cite{zhu2020}, a GNN architecture specifically designed to handle heterophily. 
H\textsubscript{2}GCN accounts for the fact that in heterophilic networks neighboring nodes may have dissimilar features and labels, unlike standard GNNs that assume instead similarity. It does so by specific design choices in the neural network architecture, such as separate $\phi_{i}^{(\ell)}$ and $\phi_{\partial i}^{(\ell)}$ for self and neighbor embeddings, skip connections and embeddings from second-order neighbors. These have been shown helping to mitigate the oversmoothing effect \cite{oversmoothing, over_gat, Oono2019GraphNN, Cai2020ANO}, which is known to worsen GNN performance in heterophilic settings.

\section{Experimental details}

\subsection{Datasets}

\paragraph{Synthetic data} 
We generate synthetic networks with community structure. We assign $K$-dimensional membership vectors to nodes, assuming an underlying community structure consisting of \(K\) overlapping groups. We then draw a \(K \times K\) affinity matrix controlling the density of edges between pairs of nodes. This follows the generative models behind the PGMs considered here. 
Specifically, we generate directed networks of $N \in\ccup{100,1000}$ number of nodes; $K=5$ communities and average degree $\langle k \rangle \in\ccup{5,20}$. To control the heterophily of the network, we manipulate the density of edges between nodes in different communities. Heterophilic have larger off-diagonal elements of the affinity matrix. For each combination of size and average degree, we generate one homophilic and one heterophilic network. We sample $10$ networks for each  parameter configuration. {More details are provided in \cref{apx:synth}}.
\vspace{-10pt}
\paragraph{Real data}
We use 10 real datasets from different domains, of varying size, sparsity, and homophily level, as reported in \cref{tbl:dataset_statistics}. Cora and Citeseer \cite{citation} are citation networks, nodes represent papers connected by citation
edges and features are bag-of-word abstracts. 
Protein-Protein Interaction (PPI) \cite{ppi} and Peptides \cite{peptides} are biological networks, edges represent interactions between proteins and similarity relationships between peptides, respectively, and features contain various protein and peptide properties.
Reddit \cite{reddit}, PolBlogs \cite{polblogs} and Email \cite{email} are information networks, describing comments between users in Reddit, links between political blogs and email communications between members of an European institution, respectively.
Chameleon, Wisconsin and Texas \cite{web_page} are web page-page networks, where edges represent hyperlinks between web pages and node features are bag-of-words representations. {Seven of these datasets contain additionally labels on nodes, these are used to estimate the level of homophily.}

\begin{table*}[ht]
\caption{Dataset statistics for the 10 real datasets analysed. $N$ is the number of nodes; $\langle k \rangle$ is the average degree of a node; $h$ is the edge homophily ratio (i.e., the fraction of edges connecting nodes with the same class label, computed only for datasets with node labels); $F$ is the dimension of the node feature (when available).}
\label{tbl:dataset_statistics}
\vskip 0.15in
\begin{center}
\begin{small}
\resizebox{\textwidth}{!}{%
\begin{tabular}{lcccccccccc}
\toprule
\textbf{Data} & Cora & Citeseer & PPI & Peptides & Reddit & PolBlogs & Email & Chameleon & Wisconsin & Texas \\
\midrule
Type & Citation & Citation & Biology & Biology & Information & Information & Information & Web & Web & Web\\
$N$ & 2708 & 3327 & 1750 & 338 & 228 & 1490 & 1005 & 2277 & 251 & 183 \\
$\langle k \rangle$ & 3.90 & 2.74 & 18.47 & 2.02 & 2.20 & 12.77 & 25.44 & 15.85 & 2.05 & 1.77 \\
$h$ & 0.81 & 0.74 & - & - & - & 0.91 & 0.36 & 0.23 & 0.21 & 0.11 \\
$F$ & 1433 & 3703 & 50 & 9 & - & - & - & 2325 & 1703 & 1703 \\
\bottomrule
\end{tabular}
}
\end{small}
\end{center}
\vskip -0.1in
\end{table*}

\subsection{Cross-validation details}

For each input network, we split the edges in training and test sets, and we employ a 5-fold cross validation with a $80\%-20\%$ split in training and test, respectively. For GNNs, we extract from the training set a validation set, corresponding to $10 \%$ of the entire dataset. We use the area under the receiver operating curve (AUC) as link prediction metric. This tells the probability that a randomly selected existing edge in the test set is assigned a higher probability of existing than a randomly selected non-existing edge. 

\paragraph{Hyperparameters tuning}
For PGMs, we cross-validate the number of communities $K$ in \mt\ and \mtcov, and $\gamma$ in \mtcov, {a parameter tuning the contribution of $A$ and $X$ in the cost function}. \bnp\ learns $K$ automatically. 
For the GNN models, we tune the following hyperparameters: learning rate $\in\ccup{0.001,0.01,0.1}$; weight-decay $\in\ccup{0.0001,0.001,0.01}$; dropout $\in\ccup{0.0,0.3,0.5}$; hidden dimension $\in\ccup{32,64,128}$; and number of layers $\in\ccup{1,2}$. Each model was trained for $200$ epochs with Adam optimizer \cite{adam} and binary cross-entropy loss. We employ an early stopping criterion and we save the weights of the model minimizing the loss on the validation set with a patience of $50$ epochs.

\section{The role of input features}
\label{expe_1}

One of the main differences between PGMs and GNNs is that the latter require node features $X$ in input, in addition to a graph adjacency matrix $A$. The choice of what features to give in input impacts performance on downstream tasks \cite{li2022expressive} and it is not clear how to make this choice in general. This is particularly relevant in datasets where there is no extra information besides the adjacency matrix, and one has to come up with dummy features to feed into the GNN. For instance, in social network analysis, user demographic data may be obscured due to privacy concerns. In such cases, synthetic features derived solely from the input graph, such as node degree and total node count, can been used as node attributes \cite{Errica2019AFC, sage, Xu2018HowPA}. Hence, our first step is to assess the role played by the nature of nodes features in learning.

For this, we run a first experiment where we change the {type of information} and the dimensionality of features given in input to GNNs and evaluate their impact on the performance of GNNs compared to that of PGMs.\\
We experiment with two kinds of features: 1) a \textit{structure-based} feature, that uses information about the edges the node is linked to; and 2) an \textit{attribute-based} feature, that uses existing node features (when these are available).
\\
In case $1)$, we assign to each node $i$ the corresponding row in the adjacency matrix $A_i = [A_{i1},\dots,A_{iN}]$. This information is readily available as it is a quantity simply derived from the input matrix $A$. We stress that, when splitting the graph in training, validation and test set {in our cross-validation link prediction routine}, we take into account the masking of the edges, and we feed the node with the masked adjacency matrix row. Indeed, the training set feature only contains the links between nodes in the training set graph, while for validation and test we progressively include the validation and test set links. \\
In case $2)$, the features are directly available only {for 7 out of 10} real datasets. {For the real datasets that do not come equipped with engineered features, we assess their performance by simply feeding the \textit{structure-based} feature.

In the case of PGMs, as input features are not necessary, we employ node features only in the model designed to take in input node features, i.e. \mtcov. {In this case we employ only one type of node feature,} one-dimensional and \textit{structure-based}. In synthetic data, the feature corresponds to the argmax of the ground truth membership vector of the node $u_i$. In real data, where we do not have  ground truth memberships, we assign to each node the cluster obtained by applying K-means to the adjacency matrix row $A_i = [A_{i1},\dots,A_{iN}]$.

\begin{figure}[ht]
    \vskip 0.2in
    \begin{center}
    \centerline{\includegraphics[width=1\textwidth]{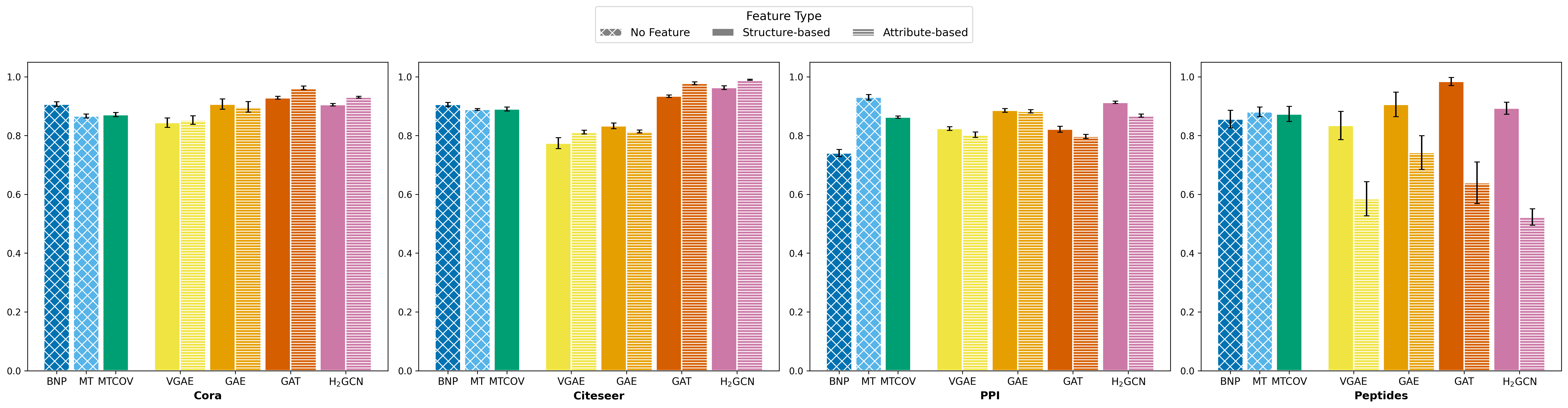}}
    \caption{{Structure-based vs. attribute-based features. AUC scores of link prediction in real datasets when giving in input different types of features. For the PGMs, only \mtcov\ uses input features. Bars and errors are mean and standard deviation over 5 cross-validation folds.}}
    \label{fig:structure-attribute}
    \end{center}
    \vskip -0.1in
\end{figure}
 
We find that GNNs' performance is generally comparable {in the two cases}, see \cref{fig:structure-attribute}, {except for Peptides, where AUC is much higher in the \textit{structure-based} case. This is also the only dataset with relatively low-dimensional features ($F=9$). Feeding the high-dimensional adjacency matrix row improves performance, which is generally comparable to that of PGMs. {We find similar results for the synthetic data, see \cref{fig:structure-ass} in Appendix \ref{add_plot}}. 
\\
This experiment shows that there are no clear advantages in utilizing an extra input feature $X$ for link prediction. Instead, one has to be careful in using all the information available, because adding a feature can even hurt performance. Similar results have been found in previous works using PGMs \cite{peel2017ground,newman2016structure,mtcovariate,badalyan2024structure}.} {We stress that prior studies \cite{cui2021positional,tenorio,TAGUCHI2021155} have proposed strategies to handle missing features in GNNs, specifically focusing on structural and positional encodings.}

\begin{figure}[ht]
    \vskip 0.2in
    \begin{center}
    \centerline{\includegraphics[width=0.85\textwidth]{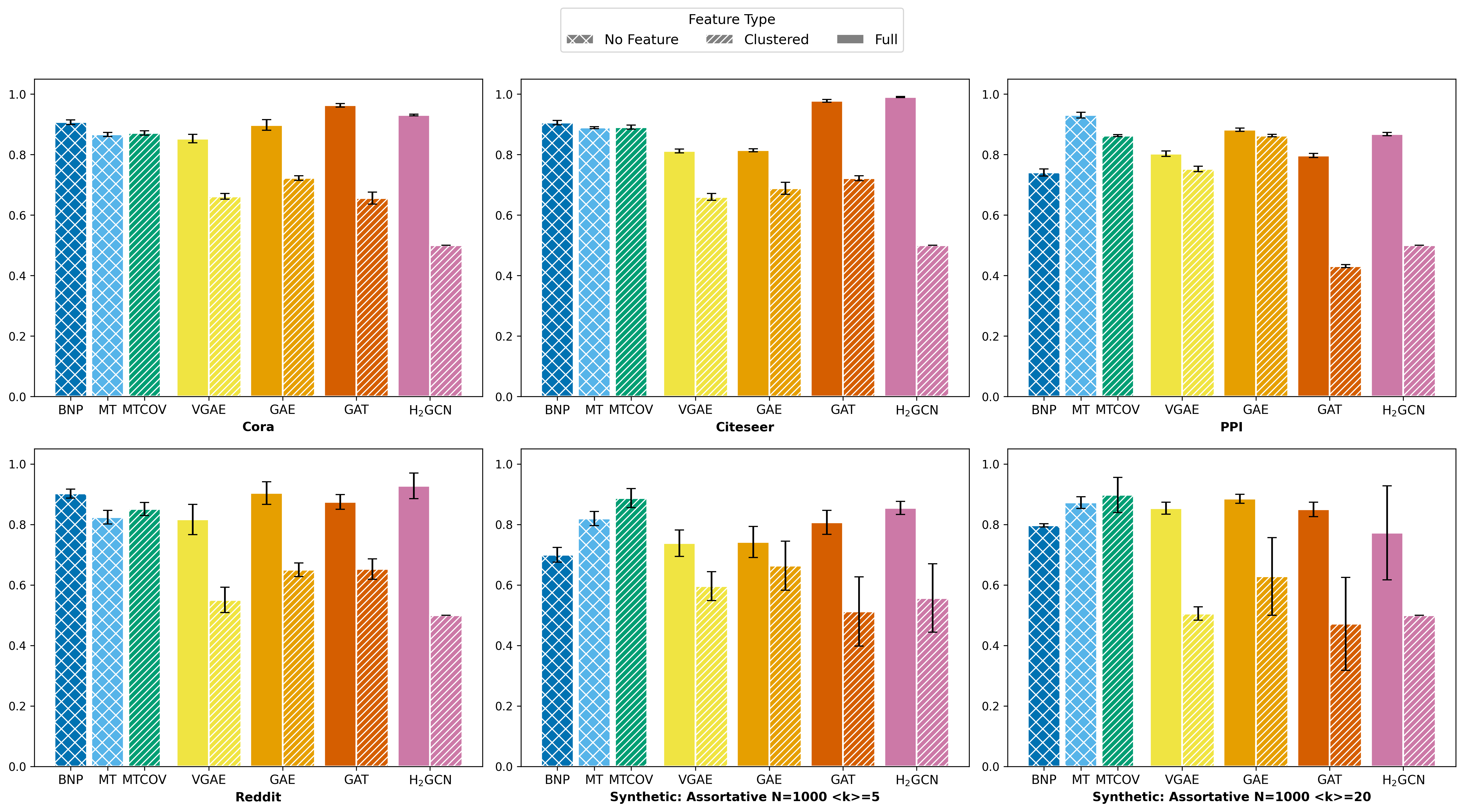}} 
    \caption{
    {High vs. low dimensional features. AUC scores of link prediction in real and synthetic datasets when giving in input either the original feature of dimension $F$ (Full) or the one clustered with K-means (Clustered), i.e. $F=1$. For the PGMs, \mtcov\ takes one-dimensional (Clustered) features by design and the other two PGMs no feature at all. Bars and errors are mean and standard deviation over 5 cross-validation folds.}}
    \label{fig:cluster}
    \end{center}
    \vskip -0.2in
\end{figure}

Motivated by the above results, we assess the impact of the feature dimensionality $F$ by reducing it to one, clustering the feature vectors with K-means and giving the cluster label as feature. This mimics situations where availability of node attributes is limited and one has access only to one type of information, e.g. age or income in a social network. We find that GNNs' performance is significantly reduced when the input feature is clustered and this happens for every dataset studied, see \cref{fig:cluster} and \cref{fig:clustered_real} in Appendix \ref{add_plot}. Instead, \mtcov, which takes in input one-dimensional features by design, shows comparable performance to that of the other PGMs and GNNs with full features. This shows a potential limitation of GNNs, the inability to either properly utilize a low-dimensional feature or to discard it in favor of, for instance, a dummy high-dimensional feature like the \textit{structure-based} one considered above. {This was also observed in a node classification task \cite{sato}, where it was shown how GNNs struggle in handling low-dimensional input features, and emphasized the benefits of graph-related features over manually engineered ones.}}

\section{Robustness to noisy features}
\label{expe_2}

In the second experiment, we investigate the robustness to noisy features  by systematically randomizing features of randomly selected nodes and observing the impact on each model's performance. This simulates scenarios where some node attributes might be missing or noisy. We perform this experiment on the real datasets equipped with \textit{attribute-based} features.

In the case of GNNs, we randomly shuffle the entries in the input vector of features ${X_i} = (X_{i1}, X_{i2}, \dots, X_{iF})$, by applying a random permutation function $\pi$ to its indices: ${{X_i}^{\text{shuffled}}} = (X_{i\pi(1)}, X_{i\pi(2)}, \dots, X_{i\pi(F)})$. {We apply different permutations $\pi$ to each node feature.} \\
In the case of \mtcov, instead, we replace the single scalar feature with a number extracted uniformly at random from the range of possible clusters, {i.e. the number of K-means clusters of the adjacency matrix row.}

We measure the impact of feature randomization on a varying percentage of nodes: $0 \%$ {(original input features are intact); $50 \%$ (features are randomized for half of the nodes); and $100 \%$ (all nodes have randomized features)}.

We find that PGMs are more robust when increasing the percentage of nodes whose features are randomized, highlighting the stronger reliance of GNNs on input feature information, see \cref{fig:noise}. In particular, GAT and H\textsubscript{2}GCN appear to be overall more sensitive to noise compared to autoencoder-based architectures (VGAE, GAE). The GAT enhanced sensitivity may be related to the fact that, when features contain noise, the attention scores can be misleading, and we notice this is especially pronounced when the input features have lower dimensionality, as in PPI ($F=50$) and Peptides ($F=9$). While VGAE and GAE generally exhibit greater resilience to noise, they become more sensitive when the dataset is heterophilic, as in Wisconsin and Chameleon.

\begin{figure}[ht]
    \begin{center}
    \centerline{\includegraphics[width=0.85\textwidth]{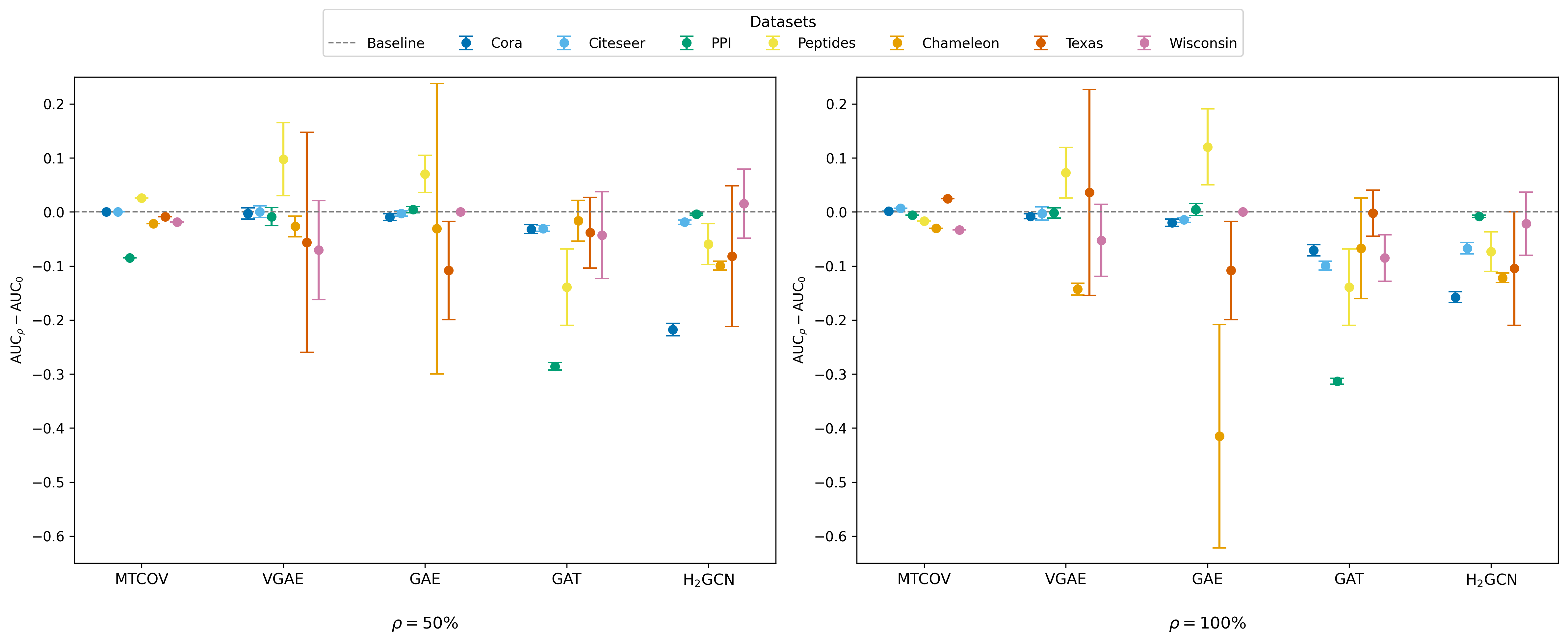}} 
    \caption{AUC score variation in real datasets when increasing the percentage of nodes $\rho$ whose \textit{attribute-based} features are randomized. The AUC score variation is calculated as the difference between the $AUC_{\rho}$ and the original $AUC_0$, corresponding to the case where $0\%$ of nodes have randomized features. Markers and errors are mean and standard deviation over 5 cross-validation folds.}
    \label{fig:noise}
    \end{center}
    \vskip -0.4in
\end{figure}

\section{Resilience to increasing heterophily}
\label{expe_3}

Having studied in detail the role of input features in learning missing links, we now focus on comparing the two frameworks for different network clustering structure. Specifically, the third experiment explores the impact of increasing the heterophily (or disassortativeness) of the graph. This is the property of observing nodes in different clusters more likely to be connected, which contrasts with homophilic (or assortative) networks, where we expect nodes in the same cluster to have higher chances of being connected. In PGMs, learning in disassortative networks has often been shown to be a more challenging task than in assortative networks \cite{multitensor,decelle2011asymptotic,ruggeri2023community,hood2025broad}. Similar declines in performance in heterophilc networks are also now increasingly investigated in GNNs \cite{luan,digiovanni,wang,edge_homo,li2022finding,luan2024heterophilic,zhu2024impact}.

In the synthetic case, we can directly manipulate the level of heterophily by varying the density of edges between nodes in different clusters in the affinity matrix \cite{multitensor}. {In real data, when extra node labels are available, and assumed to be informative, we measure the homophily edge ratio $h$ as the fraction of edges in a graph which connect nodes that have the same class label, as done e.g. in \cite{edge_homo}. A decreasing value of $h$ indicates increased heterophily. Four of the real datasets have $h<0.4$. In the synthetic case we compute $h$ by assigning each node the label corresponding to the {argmax} of its membership vector.}

\begin{figure}[ht]
    \vskip 0.2in
    \begin{center}
    \centerline{\includegraphics[width=0.85\textwidth]{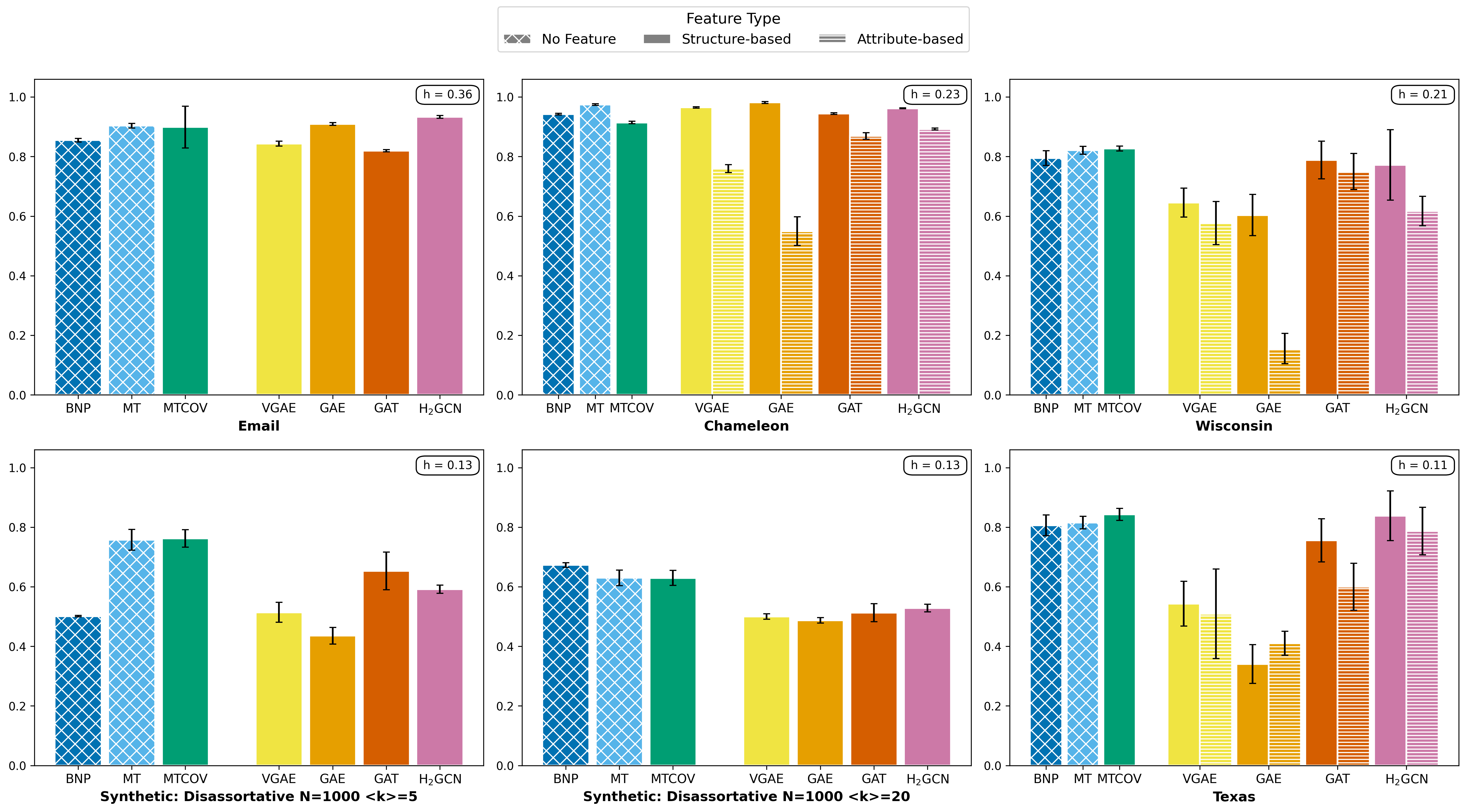}} 
    \caption{AUC scores of link prediction in real and synthetic heterophilic datasets when using the full feature. For the real dataset, we compare the AUC when feeding the GNN with \textit{structure-based} or \textit{attribute-based} features. For the PGMs, only \mtcov\ uses input features. Bars and errors are mean and standard deviation over 5 cross-validation folds.}
    \label{fig:heterophily}
    \end{center}
    \vskip -0.2in
\end{figure}

We find that PGMs generally outperform GNNs when increasing the heterophily of the graph (both synthetic and real), when given the full input feature, see \cref{fig:heterophily}. In particular, autoencoder-based architectures (VGAE, GAE) suffer a significant decrease in performance in two datasets (Wisconsin and Texas). However, we also find that by giving in input to the GNN the row of the adjacency matrix (\textit{structure-based} feature) instead of the \textit{attribute-based} feature, the situation changes favorably. In this case, GNNs show similar performance to PGMs in two datasets (Email and Chameleon).  This applies to all GNN models, including H\textsubscript{2}GCN, which is explicitly designed to handle heterophily. 
Nonetheless, even under these favorable conditions, H\textsubscript{2}GCN's performance remains comparable to or lower than that of PGMs, suggesting that architectural adaptations alone may not fully address the challenges posed by heterophilic graphs. This result highlights that while GNNs are expected to be generally less robust in heteropilic graphs, performance could be improved by a better choice of input features, rather than acting on a change in architecture design as proposed in \cite{digiovanni,zhu2020,difrancesco}. A similar conclusion was also observed in node classification tasks \cite{ma2021homophily}, where GNN architectures were shown to significantly improve their performance under good conditions of heterophily, specifically when nodes sharing the same label also exhibit similar embeddings and neighbor distributions. Moreover, the phenomenon of oversmoothing \cite{oversmoothing, over_gat, Oono2019GraphNN, Cai2020ANO}, which is commonly associated with poor performance on heterophilic graphs \cite{digiovanni,ma2021homophily,ordered_gnn}, has been shown to be avoidable without altering the GNN architecture, simply by increasing the variance of weight initialization~\cite{gnns_do_not_over}.

\section{Interpretability and complexity}

\subsection{Interpretability}
\label{interpretability}

PGMs and GNNs are inherently different in terms of their interpretability.
The PGMs analysed here learn community membership vectors on each node. These could be either hard or soft, depending on whether nodes are allowed to belong to more than one group. Memberships are quantities directly interpretable as they provide clear insights into the community structure of the network. In addition, the linearity of these models allows a direct mapping between these parameters and the probability of observing an edge between two nodes. This is usually a design choice driven by mechanistic principles.\\
Unlike PGMs, GNNs learn embedding vectors, quantities that are not directly interpretable. Instead, one has to make ad-hoc processing a posteriori to allow for interpretability \cite{li2022interpretable} {or design GNN models specifically to optimize for it \cite{huang2022graphlime}}. A common choice is to cluster them with K-means \cite{Mehta2019StochasticBM, tsitsulin2023graph, tian, attributed,shin}, thus assigning them one-dimensional labels. {This is a simple and direct way to analyze GNN outputs and matches our approach of prioritizing general-purpose GNN designs. Alternative and more advanced interpretability techniques exist, but they require modifying the neural network architecture itself or the loss function employed \cite{attributed,shin}. Hence, we do not explore them here.} Similarly, it is not directly clear how to combine node embeddings to estimate the probability of an edge. One has to choose a function that combines pairs of embedding, whose output is then associated to edges. Possible choices are the dot product or a concatenation \cite{paire, automated}.

{To create a parallelism with the interpretation of communities in  PGMs and the output GNNs' embeddings, we visualize the memberships and embeddings in a two-dimensional plan by applying a dimensionality reduction to these vectors using t-SNE \cite{tsne}. We use a synthetic network as example, as we have access to ground truth (GT) communities. We first notice how nodes that have a more pronounced mixed-membership are placed on the border of their main community, closer to the other communities they also participate in, see \cref{fig:tsne}. This is somehow expected when plotting the GT $u_i$, given their shape. However, it is not trivial that the models are able to maintain this positional property in the inferred parameters. We find that both \mt\ and GAT are able to place more mixed nodes closer to the border, as observed for the GT. For instance, nodes that belong to a large extent to two communities, are placed in between the two main clusters. However, while mixed-membership PGMs allow for a direct estimation of the mixed-membership, as this is their natural output, it is not clear how to extract this information automatically and in a principled way {(beside visual inspection) in GNNs from their output embeddings, as they lack the innate ability to represent these memberships explicitly.}
To enhance the interpretability of GNNs, one promising direction is to develop methods that enable the extraction of interpretable mixed-membership information from their embeddings. Achieving this, however, would require architectural modifications to GNNs, as done in \cite{Mehta2019StochasticBM}.

\begin{figure*}[ht]
    \vskip 0.2in
    \begin{center}
    \centerline{\includegraphics[width=1\textwidth]{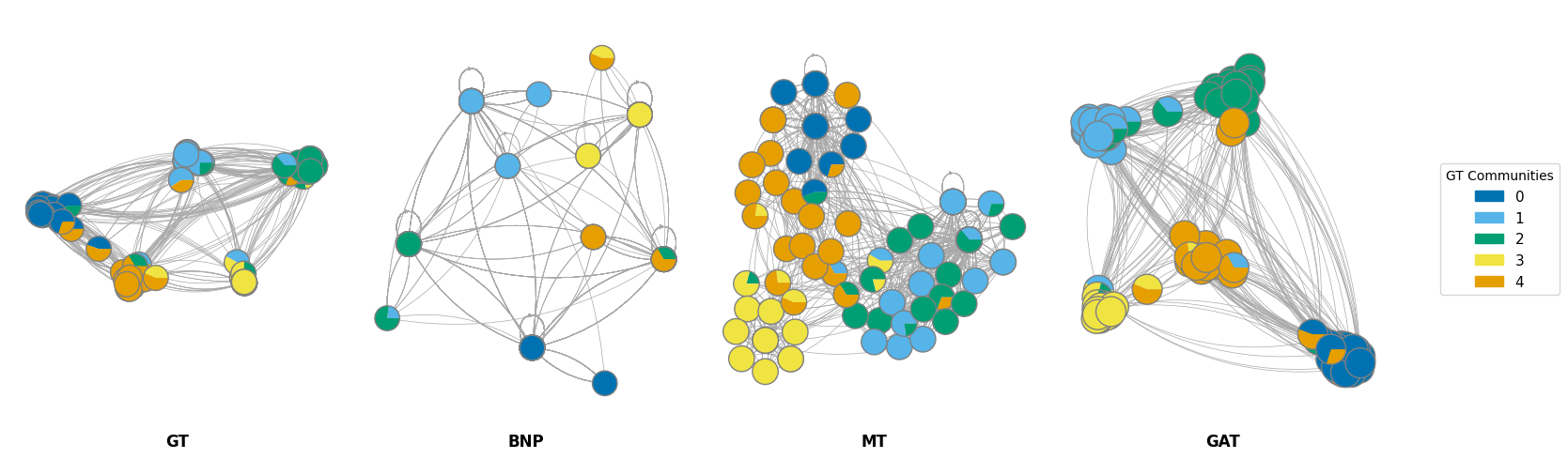}}
    \caption{{Comparing the parameters inferred. Marker positions are extracted by applying t-SNE to either the memberships $u_i$ (PGMs) or the embeddings $h_i^{(L)}$ (GNNs), and we compare with the ones in ground truth (GT). We consider an example synthetic network, with $N=100$ and $
    \langle k \rangle=20$. Colors represent GT mixed-memberships. All the AUC values are between $0.85$ and $0.88$.}}
    \label{fig:tsne}
    \end{center}
    \vskip -0.2in
\end{figure*}

{Furthermore, we see how models favor different clusters in real data, for a similar model expressiveness, measured by similar AUC values. We show an example of this for the Citeseer dataset in \cref{fig:membership}, where we plot the inferred parameters. In this example, \bnp\ tends to place nodes in more and smaller clusters, \mt\ and \mtcov\ have similar result (we report only one), with 7 clusters with low amount of mixed-membership, and GAT has one dominating larger cluster. All the AUC values are between $0.89$ and $0.93$, indicating that the models are reconstructing $A$ well, but they do so by learning different partitions.}

\begin{figure}[ht]
    \vskip 0.2in
    \begin{center}
    \centerline{\includegraphics[width=0.65\textwidth]{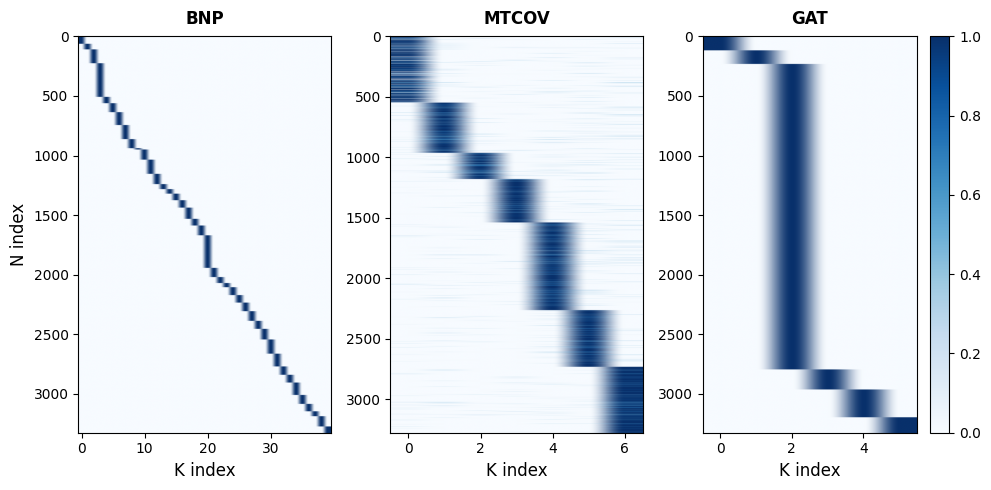}}
    \caption{Inferred partitions on Citeseer dataset. For the PGMs we plot the membership vectors, mixed for MTCOV and hard for BNP. For GAT, we first cluster the final embeddings using K-Means with the number of clusters set to 6 (matching the number of classes), then we generate hard-membership vectors based on the clustering results. The nodes are ordered depending on the community assignment \( K \).}
    \label{fig:membership}
    \end{center}
    \vskip -0.2in
\end{figure}

\subsection{Computational Complexity}
\label{complexity}

In addition to interpretability, PGMs and GNNs also differ in their computational complexity, as shown in \cref{comp_complexity}.
We notice in particular that all the models are linear (or almost, in \bnp) in the system size (the number of nodes or edges). GNNs have the number of layers $L$ and the embedding dimensions $F'$ as multiplicative factors, while PGMs have the number of communities $K$ (at most squared). Usually, $L$ is kept small, to prevent issues such as oversmoothing \cite{oversmoothing, over_gat, Oono2019GraphNN, Cai2020ANO}. Hence, we conclude that they all have comparable performance.

\begin{table}[ht]
\caption{{Computational complexities for the models under analysis. \( N \) the number of nodes, \( E \) the total number of edges, \( E^{(2)} \) the number of edges between nodes that are connected via a 2-hop path, \( d_{\max} \) the maximum degree across all nodes, \( K \) the number of communities, \( Z \) the number of attribute categories, \( F \) the input feature dimension, \( F' \) the hidden layer dimension, and \( L \) the number of GNN layers.}}
\label{comp_complexity}
\vskip 0.15in
\begin{center}
\begin{small}
\begin{sc}
\begin{tabular}{lllcr}
\toprule
Class &Model  & Computational Complexity \\
\midrule
PGM  &MT & \(O(E \cdot K^2) \) \\
 &MTCOV & \( O(E \cdot K^2 + N \cdot K \cdot Z) \) \\
 &BNP &  \( O(N\ln(N^2)) \) \\
 \specialrule{0.5pt}{2pt}{3pt}
GNN &GAE & \(O(L \cdot (E \cdot F' + N \cdot F \cdot F'))\) \\  
 &VGAE &   \(O(L \cdot (E \cdot F' + N \cdot F \cdot F'))\) \\ 
 &GAT & \(O(L \cdot ( E \cdot F' + N \cdot F \cdot F'))\)\\ 
& H$_2$GCN & \(O(E \cdot d_{\max} +  L \cdot ( (E + E^{(2)}) \cdot F') + N \cdot F \cdot F')\) \\
\bottomrule
\end{tabular}
\end{sc}
\end{small}
\end{center}
\vskip -0.1in
\end{table}

\section{Conclusions}
\label{concl}

In this work, we compare Probabilistic Graphical Models (PGMs) and Graph Neural Networks (GNNs) in how they handle network data, and we conduct our experiments on a link prediction task. 

Our findings reveal that PGMs outperform GNNs when input features are either low-dimensional or noisy, a common scenario in real-world networks where node attributes may be minimal or unreliable. Additionally, we observe that PGMs exhibit greater robustness to increasing heterophily of the graph compared to GNNs. Interestingly we find that GNNs performance on heterophilic graphs could be improved by a better choice of input features, as similarly shown for a node classification task in \cite{tsitsulin2023graph,ma2021homophily}.

{However, an important advantage of GNNs lies in their flexibility in processing input features. Unlike PGMs, GNNs do not rely on ad-hoc design choices to handle heterogeneous features, thus simplifying their implementation and widening their usage. Nevertheless, recent work \cite{10.1093/pnasnexus/pgaf005} suggests that it may be possible to benefit from similar black-box tools as those used in GNNs to incorporate arbitrary types of attributes as input to PGMs.}

Moreover, previous studies have shown that linear models can outperform GNNs in link prediction tasks \cite{wang2022powerful,srinivasan2019equivalence}. Given these limitations, GNNs should ideally be designed to learn automatically whether to discard irrelevant features or to select what features to use, potentially incorporating structural information such as the adjacency matrix of the graph. In the absence of this additional learning step, care should be taken when choosing what features to give in input. 

\section*{Acknowledgements}

This research project has been developed during a visiting period at Max Planck Research School for Intelligent Systems (IMPRS-IS), Tübingen, Germany.


\newpage
\appendix
\onecolumn

\section{Additional plots}
\label{add_plot}

In this section, we present two additional plots complementing the input features experiments {presented} in the main text: i) in \cref{fig:structure-ass} we compare the performance of PGM and GNN models when using \textit{structure-based} input features on both real and synthetic datasets; ii) in \cref{fig:clustered_real} we further illustrate the performance of GNNs when fed either the full input feature or the clustered one on both real and synthetic datasets. 

As shown before in \cref{fig:structure-attribute}, from \cref{fig:structure-ass} we can see that the performance of GNNs is generally comparable to that of PGMs when feeding \textit{structure-based} input features.

\begin{figure}[htb]
    \centering
    \includegraphics[width=0.75\textwidth]{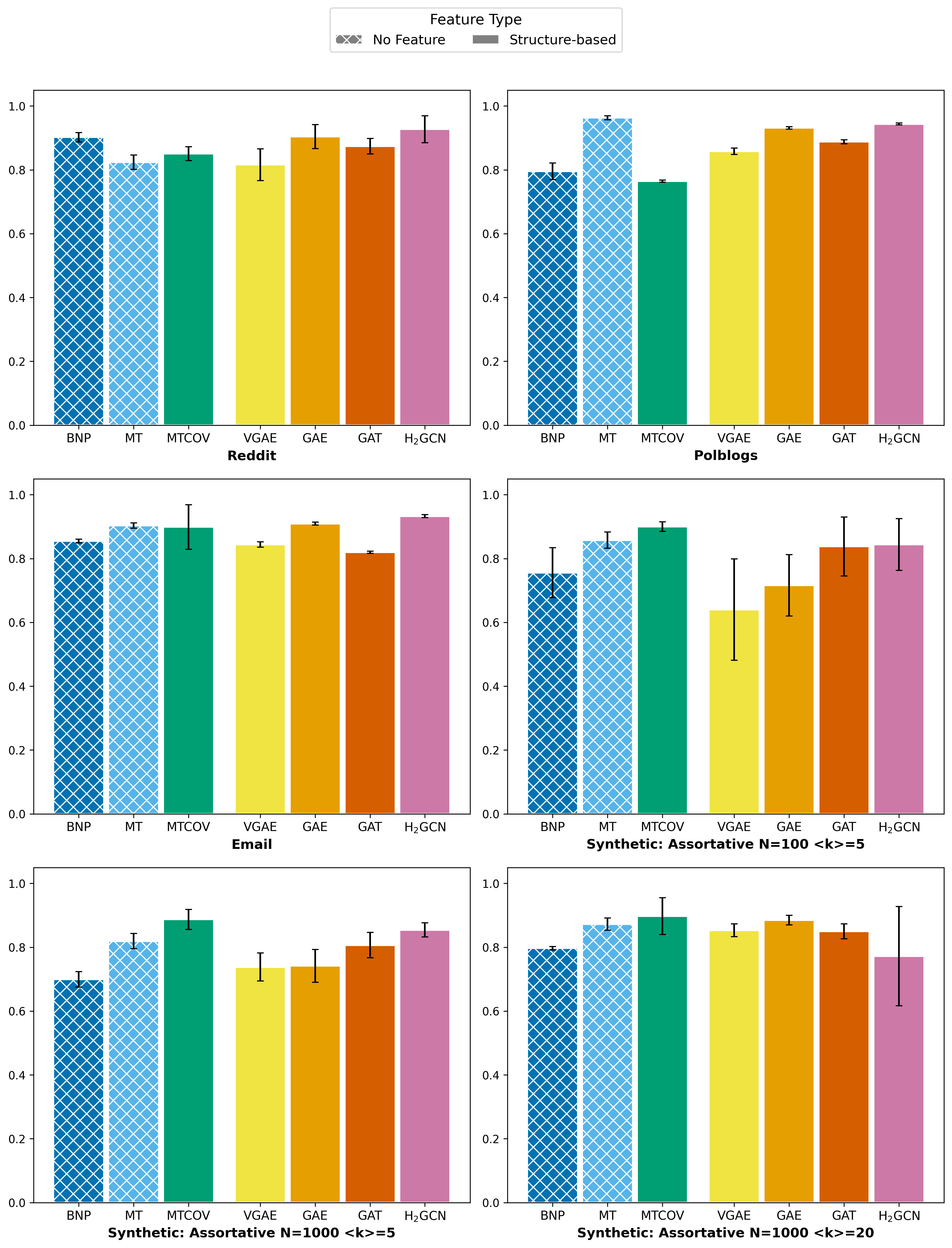}
    \caption{AUC scores of link prediction in real and synthetic datasets when using \textit{structure-based} input features. For the PGMs, only \mtcov\ uses input features. Bars and errors are mean and standard deviation over 5 cross-validation folds.}
    \label{fig:structure-ass}
\end{figure}

\newpage

Furthermore, as shown previously in \cref{fig:cluster}, we observe a significant drop in GNN performance when the input features are clustered, and this effect is consistent across most datasets. The difference in performance between full and clustered features tends to be smaller when the input is already low-dimensional, as in the case of Peptides ($F=9$), or when the full feature performance on the dataset is already low, such as in the disassortative synthetic setting.

\begin{figure}[htb]
    \centering
    \includegraphics[width=1\textwidth]{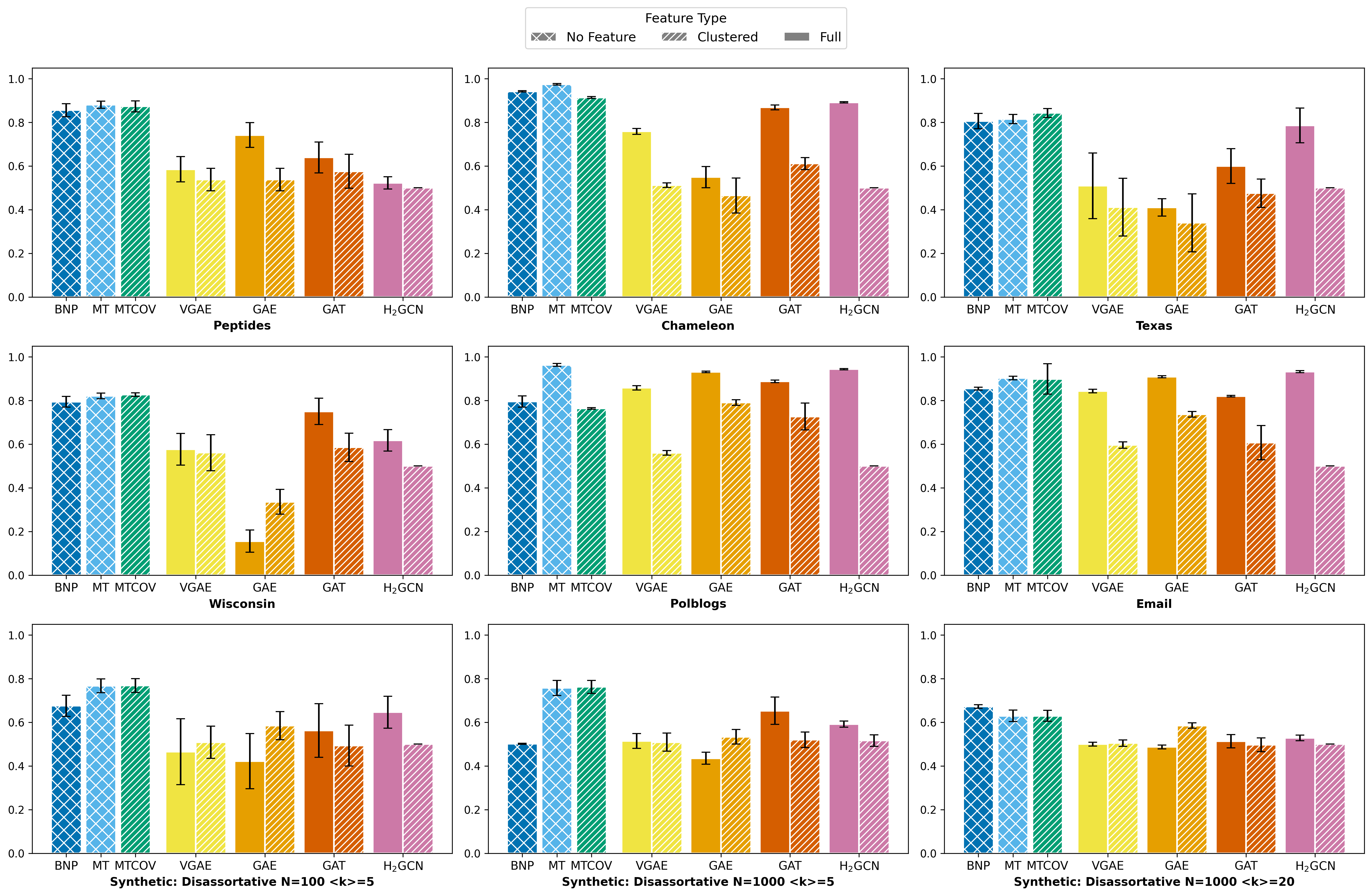}
    \caption{High vs. low dimensional features. AUC scores of link prediction in real and synthetic datasets when giving in input either the original feature of dimension $F$ (Full) or the one clustered with K-means (Clustered), i.e. $F=1$. For the PGMs, \mtcov\ takes one-dimensional (Clustered) features by design and the other two PGMs no feature at all. Bars and errors are mean and standard deviation over 5 cross-validation folds.}
    \label{fig:clustered_real}
\end{figure}

\newpage
\section{Synthetic dataset details}
\label{apx:synth}

We generate the synthetic networks following the generative modeling provided by \mt\ \cite{multitensor}. The network consists of \(N\) nodes and its connectivity is described by an adjacency matrix \(A\), where \(A_{ij}\) represents the number of edges from node \(i\) to node \(j\). The networks are generated probabilistically, assuming an underlying structure consisting of \(K\) overlapping groups. Each node belongs to each group to an extent described by a \(K\)-dimensional membership vector. In order to generate directed networks, each node \(i\) has two membership vectors, \({u}_i\) and \({v}_i\), which determine how \(i\) forms outgoing and incoming links, respectively. When modeling undirected networks, we set \({u}_i = {v}_i\). {Then, the density of edges within and across communities is regulated by a $K \times K$-dimensional affinity matrix \({w}\).\\
Given the parameters ${u}_i, {v}_j, {w}$, we sample entries $A_{ij}$ of the adjacency matrix from a Poisson distribution as:}

\begin{align}
\label{eqn:bilinear}
A_{ij} &\sim \text{Pois}(M_{ij}) \\
M_{ij} &= \sum_{k, \ell=1}^{K} u_{ik} v_{j\ell} w_{k\ell} \, .
\end{align}

In our current applications, we concentrate on the sparse case, where \( M_{ij} \) is small, and we then clip \( A_{ij} \) to consider only binary entries $A_{ij} \in \ccup{0,1}$.

To generate networks with either assortative or disassortative community structure, we control the form of the affinity matrix \({w}\), which governs the interaction strengths between latent communities. In the assortative setting, \({w}\) is generated as a one-dimensional array of size \( K \), corresponding to a diagonal affinity matrix. This implies that nodes are more likely to connect within their own community, resulting in higher intra-community edge density. In contrast, for disassortative networks, \({w}\) is sampled as a full \( K \times K \) matrix, where the off-diagonal entries enable significant interactions between different communities. The entries of \({w}\) are independently drawn from a Gamma distribution with shape and rate parameters both equal to 1.0, i.e. $w_{kq} \sim \text{Gamma}(1.0, 1.0)$.  Node community memberships \({u}_i \) and \({v}_j \) are drawn from a symmetric Dirichlet distribution ${u}_i, {v}_j \sim \text{Dirichlet}(\alpha \mathbf{1})$, with $\alpha = 0.05$ ensuring that each node predominantly belongs to a small subset of communities; {and $\mathbf{1}$ being a $K$-dimensional vector with entries equal to $1$.}

To summarize, the parameters of the SBM are generated as:

\begin{align}
    {u}_i, {v}_i &\sim \text{Dirichlet}(\alpha \mathbf{1}) \ \text{with } \alpha = 0.05, \quad \text{for each node } i = 1, \ldots, N \\
    w_{k\ell} &\sim 
    \begin{cases}
        \text{Gamma}(a, b) & \text{for all } k, \ell \in \{1, \ldots, K\} \quad \text{(disassortative)} \\
        \text{Gamma}(a, b) \cdot \delta_{k\ell} & \text{(assortative)} \quad,\\
    \end{cases}
\end{align}

where $\delta_{k\ell}$ is the Kronecker delta ensuring a diagonal matrix in the assortative case and $a = b = 1.0$ in both cases.

We generate directed networks of $N \in\ccup{100,1000}$ number of nodes; $K=5$ communities and average degree $\langle k \rangle \in\ccup{5,20}$. For each combination of size and average degree, we generate one homophilic and one heterophilic network. We sample $10$ networks for each parameter configuration.

\end{document}